\documentclass[10pt,twocolumn,letterpaper]{article}

\usepackage{cvpr}
\usepackage{times}
\usepackage{epsfig}
\usepackage{graphicx}
\usepackage{amsmath}
\usepackage{amssymb}
\usepackage{subcaption}

\usepackage{bm}

\usepackage{enumitem}
\usepackage{booktabs}
\usepackage{algpseudocode,algorithm,algorithmicx}

\usepackage[pagebackref=true,breaklinks=true,letterpaper=true,colorlinks,bookmarks=false]{hyperref}

\cvprfinalcopy % *** Uncomment this line for the final submission

\def\cvprPaperID{37} % *** Enter the CVPR Paper ID here

\ifcvprfinal\fi
\begin{document}

%%%%%%%%% TITLE
\title{Evolving Losses for Unlabeled Video Representation Learning}

\author{
AJ Piergiovanni, Anelia Angelova, Michael S. Ryoo \\
  Google Brain\\
  \texttt{\small\{ajpiergi,anelia,mryoo\}@google.com} \\
}

\maketitle
%\thispagestyle{empty}

%%%%%%%%% ABSTRACT
%\begin{abstract}

%We present a new method to learn video representations from unlabeled data. Given large-scale unlabeled video data, the objective is to benefit from such data by learning a generic and transferable representation space that can be directly used for a new task such as zero/few-shot learning. We formulate our unsupervised representation learning as a multi-modal, multi-task learning problem, where the representations are also shared across different modalities via distillation. Further, we also introduce the concept of finding a better loss function to train such multi-task multi-modal representation space using an evolutionary algorithm; our method automatically searches over different combinations of loss functions capturing multiple (self-supervised) tasks and modalities.
%Our formulation allows for the distillation of audio, optical flow and temporal information into a single, RGB-based convolutional neural network. We also compare the effects of using additional unlabeled video data and evaluate our representation learning on standard public video datasets.

%\end{abstract}

%and perform an evolutionary search over the loss function capturing all the tasks and modalities.

%%%%%%%%% BODY TEXT
\section{Introduction}

We present a new method to learn video representations from large-scale unlabeled video data. We formulate our unsupervised representation learning as a multi-modal, multi-task learning problem, where the representations are also shared across different modalities via distillation. Our formulation allows for the distillation of audio, optical flow and temporal information into a single, RGB-based convolutional neural network. We also compare the effects of using additional unlabeled video data and evaluate our representation learning on standard public video datasets.

We newly introduce the concept of using an evolutionary algorithm to obtain a better multi-modal, multi-task loss function to train the network. AutoML has successfully been applied to architecture search and data augmentation. Here we extend the concept of AutoML to unsupervised representation learning by automatically finding the optimal weighting of tasks for representation learning.

\paragraph{Related Works} There are many tasks for self-supervised learning such as predicting if frames appear in order \cite{misra2016shuffle, lee2017unsupervised,pickup2014seeing}, reconstruction or prediction of future frames \cite{srivastava2015unsupervised}, time-contrastive learning \cite{timecontrastive,sermanet2017time}. Others leaned representations taking advantage of audio and video features by predicting if an audio clip is from a video or not \cite{arandjelovic2017look} or if an audio sample is temporally aliened with a video clip \cite{korbar2018cooperative}. Multi-task self-supervised learning has also shown promising results \cite{doersch2017multi}, however it assumes all self-supervised tasks have equal weightings.

%Our goal is to find a signal (i.e., combination of tasks) that can replicate supervised training without large-scale labeled datasets. Our main contributions are:

%\begin{itemize}[nosep,leftmargin=*]
%  \item Formulation of unsupervised learning as multi-modal, multi-task learning
%  \item Use of distillation to transfer features from different modalities into a single-stream network, allowing for faster computation while still capturing multi-modal features.
%  \item Evolutionary search for a loss function that automatically discovers the tasks that are beneficial for unsupervised representation learning.
%  \item Find a powerful video representation that matches performance of networks trained on supervised data (e.g., ImageNet).
%\end{itemize}

\section{Method}

\begin{figure*}
    \centering
    \includegraphics[width=0.9\linewidth]{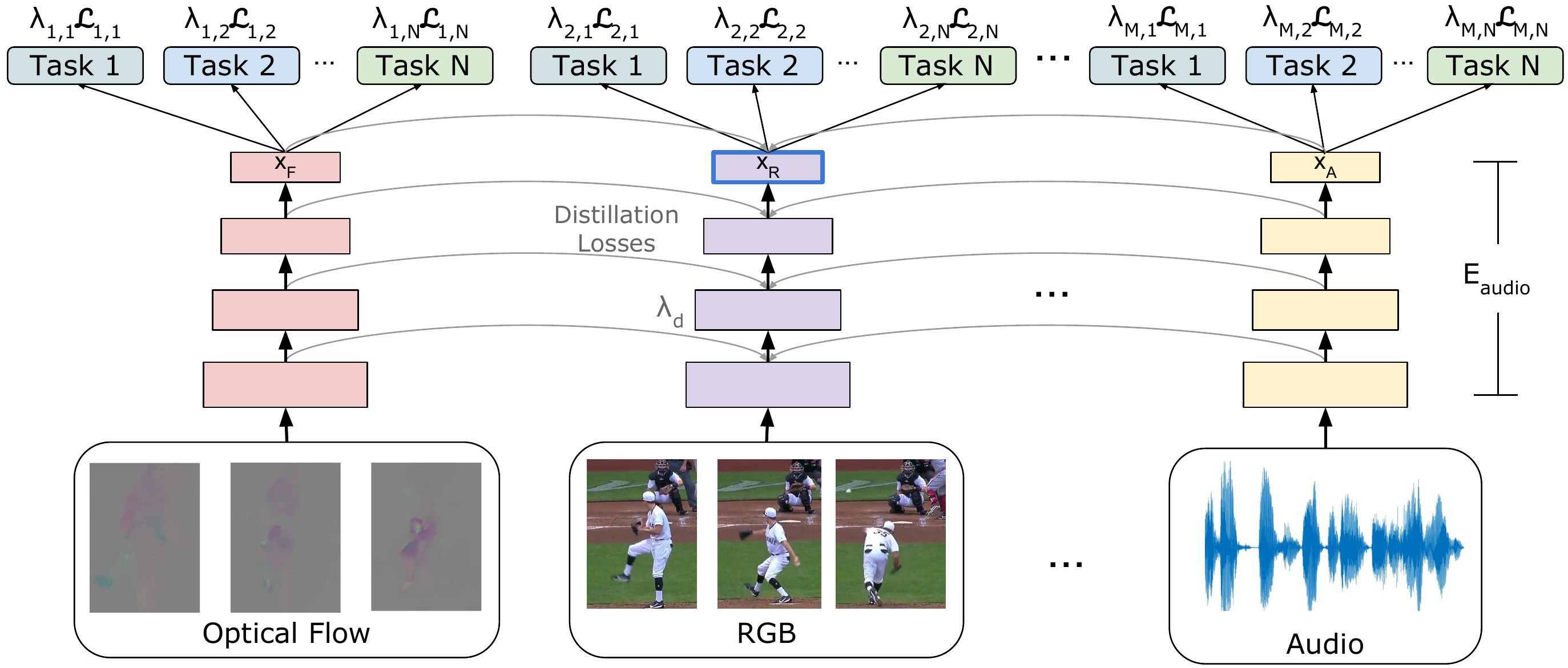}
    \caption{Overview of our multi-task, multi-modal, unsupervised representation learning framework. Each modality is trained to optimize a set of tasks. Distillation regularization loss terms `infuse' each modality's information into the main RGB network (drawn center). We evolve the loss function to automatically find optimal weights for each task and distillation location. The goal is to obtain representation from the RGB network that transfer to recognition tasks.}
    %\todo{It'll be helpful to say that the goal is unsup feat. representation - which the main net will "carry"}
    \label{fig:overview}
\end{figure*}

We formulate our video representation learning using unlabeled data as a combination of multi-task, multi-modal learning. The objective is not only to take advantage of multiple self-supervised tasks for the learning of a (good) representation/embedding space, but also to do so across multiple modalities. The idea is that synchronized multi-modal data sharing the same semantic content could benefit representation learning of the others, implemented with `distillation' losses. Fig. \ref{fig:overview} illustrates our overall model. %To better take advantage of the multi-modal representations, we use distillation to `infuse' the other modalities into the RGB network at different locations. Our final objective is to train a single RGB network that provides a strong representation for video understanding. Our formulation allows the RGB network to learn representations from various tasks and modalities.

\begin{figure}
    \centering
    \includegraphics[width=\linewidth]{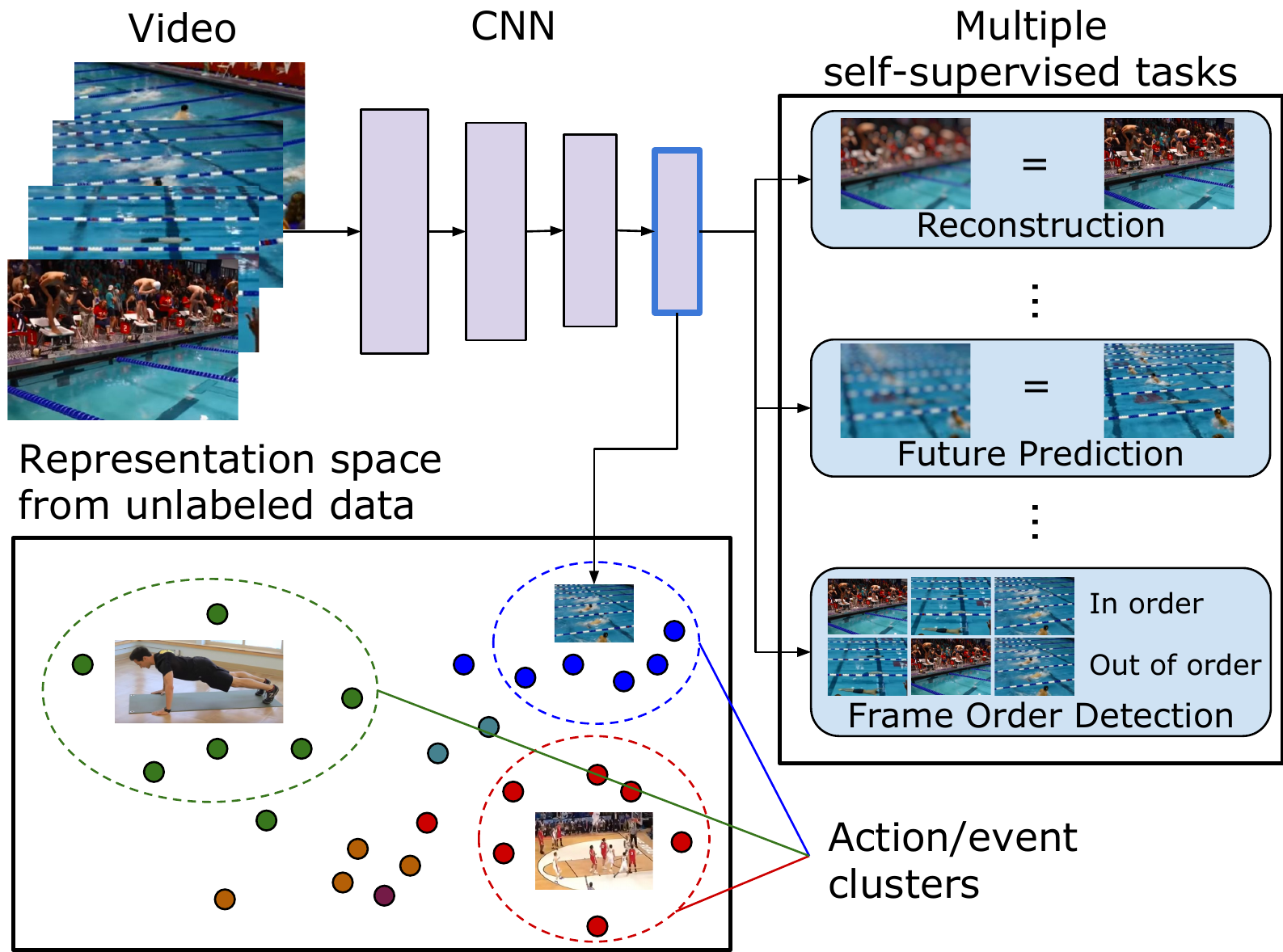}
    \caption{Overview of our multi-task, self-supervised learning framework. The objective is to obtain a good representation (a blue boundary box) from a set of self-supervised tasks. We train a single network on a variety of tasks for each input modality (note, here we only illustrate one modality/network). We use an evolutionary algorithm to automatically find the optimal combination of tasks.}
    \label{fig:concept}
    \vspace{-5mm}
\end{figure}

Importantly, we introduce the new concept of automatically \emph{evolving} the loss function. Certain tasks and modalities are more relevant to the final task so the representation needs to focus on those more the others. The idea is to search for how different multi-task and distillation losses should be combined (instead of hand-crafting a loss function with trial-and-error).

We consider many tasks each having their own loss functions. Let ${L}_{m,t}$ be the loss from task $t$ and modality $m$. We combine the multi-task learning losses during unsupervised training by weighted sum and combine it with a number of distillation losses $\mathcal{L}_d$ which fuse multiple modalities:
\begin{equation}
    \mathcal{L} = \sum_m \sum_t \lambda_{m,t} \mathcal{L}_{m,t} + \sum_d \lambda_d \mathcal{L}_d
    \label{eq:main}
\end{equation}
where $\lambda_{m,t}$ and $\lambda_d$
are the weights for the specific losses. The weight sum, $\mathcal{L}$, is the loss we use to train the model.

\begin{figure*}
    \centering
    \includegraphics[width=0.9\linewidth]{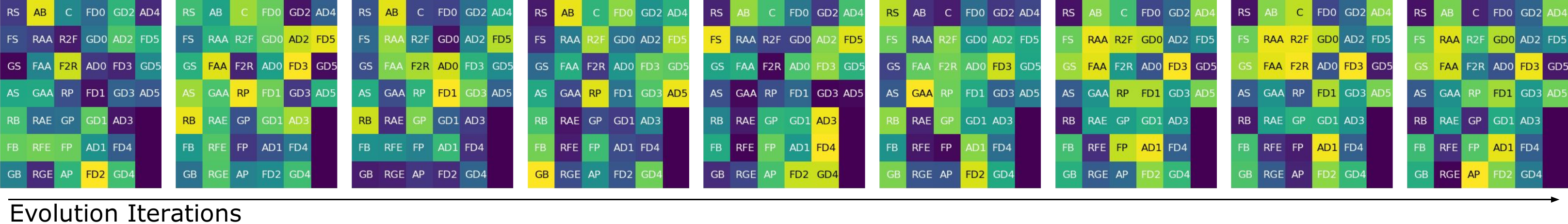}
    \caption{Evolution of the weights deciding our final loss function. Each square represents a $\lambda_{m,t}$ and how it changes over the evolutionary search. For more detailed explanation of what each weight symbol means, please check Fig.~\ref{fig:loss-heatmap}.}
    \label{fig:loss-fn-evol}
    \vspace{-5mm}
\end{figure*}

\vspace{-1em}

\paragraph{Distillation}
Distillation was introduced to train smaller networks by matching representation of deeper ones, and thus maintaining performance. Here, we use distillation to `infuse' representations of different modalities into the main, RGB network. Note that we distill representations jointly while training, whereas previous works used distillation from pre-trained networks to the networks being trained. The distillation loss is the $L_2$ difference between the activations of a layer in the main network ($M_i$) and a layer in an other network/modality ($L_i$). This encourages the activations of the main network to match the activations of other modalities, infusing other features into the main network.

\subsection{Evolving loss function}

Instead of hand-crafting the loss function, we use an evolutionary algorithm to determine the optimal weightings in Eq.~\ref{eq:main}. The weightings reflect the importance or relevance of each task and modality on the main task. Our search space consists of all the weights of the loss function. Each $\lambda$ is constrained to be in $[0,1]$. Our evolutionary algorithm maintains a population where each individual is a set of weight values the compose the final loss function. Initially, the population is random weights, uniformly sampled from $[0,1]$. At each round of evolution, a top-performing individual is chosen and one value is randomly changed.

In order to measure the fitness of each individual, we apply a clustering algorithm on the representation learned with the loss function. We first train the network using random, unlabeled videos with its loss function. Next, a subset of the HMDB training set was used for the clustering, and its similarity to the actual class clusters is measured as the fitness of the individual.

\vspace{-4mm}
\paragraph{Self-supervised tasks}
\label{subsec:tasks}

\begin{figure}
    \centering
    \includegraphics[width=\linewidth]{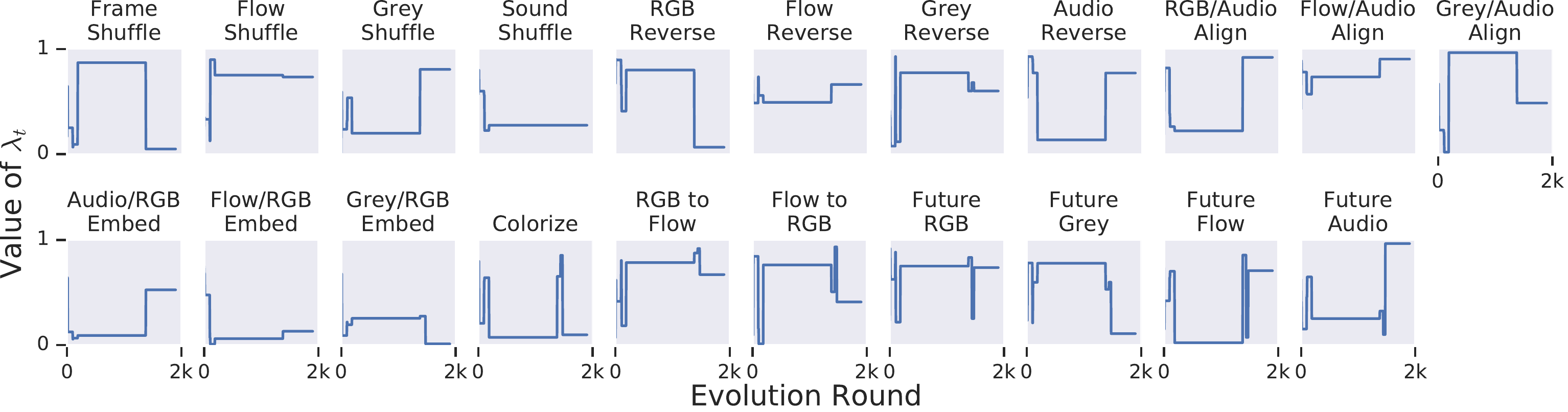}
    \caption{The values of the loss function for the various tasks throughout evolution. Higher weight values indicate the task is more important. The learned loss function automatically finds the tasks that most benefit recognition. For example, RGB frame shuffle, colorization, and joint grey/RGB embedding tasks are not too useful while the aligning the various modalities and predicting flow are.}
    \label{fig:tasks-evol}
    %\vspace{-5mm}
\end{figure}

Many tasks have been designed for unsupervised learning and we let the evolved loss function automatically discover which are important and the optimal relative weightings. Some tasks are frame reconstruction or prediction \cite{srivastava2015unsupervised}, audio-video alignment \cite{korbar2018cooperative}, generating flow from RGB inputs, etc. All tasks are listed in Fig. \ref{fig:tasks-evol}.

\begin{table}
\small
    \centering
    \begin{tabular}{lccc}
    \toprule
       Method  & $k$-means & 1-layer & fine-tune \\
    \midrule
        \multicolumn{4}{l}{\textbf{Supervised using additional labeled data}}\\
        Scratch (No Pretraining) & 15.7 & 17.8 & 35.2 \\
        ImageNet Pretrained & 32.5 & 37.8 & 49.8 \\
        Kinetics Pretrained & 68.8 & 71.5 & 74.3 \\
        \midrule
        \multicolumn{4}{l}{\textbf{Unsupervised using unlabeled videos}}\\
        Frame Shuffle \cite{misra2016shuffle} & 22.3 & 24.3 & 28.4 \\
        Reverse Detection & 21.3 & 24.3 & 27.5\\
        Audio/RGB Align \cite{korbar2018cooperative}     & 32.4 & 36.8 & 40.2 \\
        RGB to Flow   & 31.5 & 36.4 & 39.9 \\
        Predicting 4 future frames & 31.8 &  35.8 & 39.2 \\
        Joint Embedding & 29.4 & 32.5 & 38.4 \\
        \midrule
        \multicolumn{4}{l}{\textbf{Ours using unlabeled videos}}\\
        Random Loss & 25.4 & 27.6 & 30.4\\
        Evolved Loss & 44.2 & 62.8 & 66.2 \\
    \bottomrule
    \end{tabular}
    \caption{Evaluation of various self-supervised methods on HMDB51. We compare to a randomly initialized, ImageNet pretrained and Kinetics pretrained networks. We also compare to various single-task baselines and a loss function randomly sampled from our search space. All tasks were trained on our unlabeled videos.}
    \label{tab:hmdb-1}
\end{table}

\section{Experiments}

\paragraph{Evaluation of learned representations} We evaluate the representations in 3 settings: (1) $k$-means clustering of the representations (2) fixing the weights of the network and training a single, fully-connected layer for classification and (3) fine-tuning the entire network.

%We use a 3d ResNet-50 as our backbone networks. Given a loss function, we train the network for 250 epochs on unlabeled data. The learning rate is set to 0.1 (during both the evolution and the final training). We use cosine learning rate decay with a warmup period of 2 epochs.

\paragraph{Data} In contrast to previous works, which use videos from existing datasets, we use random, unlabeled YouTube video clips. Previous works used videos from datasets (e.g., Kinetics or AudioSet in \cite{korbar2018cooperative}) and simply discarded the labels. However, it is not learning from truly unlabeled data, as the samples in those datasets are biased -- they have been selected to belong to certain classes and video clips are further trimmed to regions with activities occurring.

Our dataset is truly random videos taken from YouTube and random 10-second clips are extracted. We use no labels, no data cleaning, and no human verification of any videos. This makes our data both more challenging and realistic for evaluation of unsupervised learning methods. Our dataset consists of 2 million clips.

\begin{figure}
    \centering
    \includegraphics[width=0.8\linewidth]{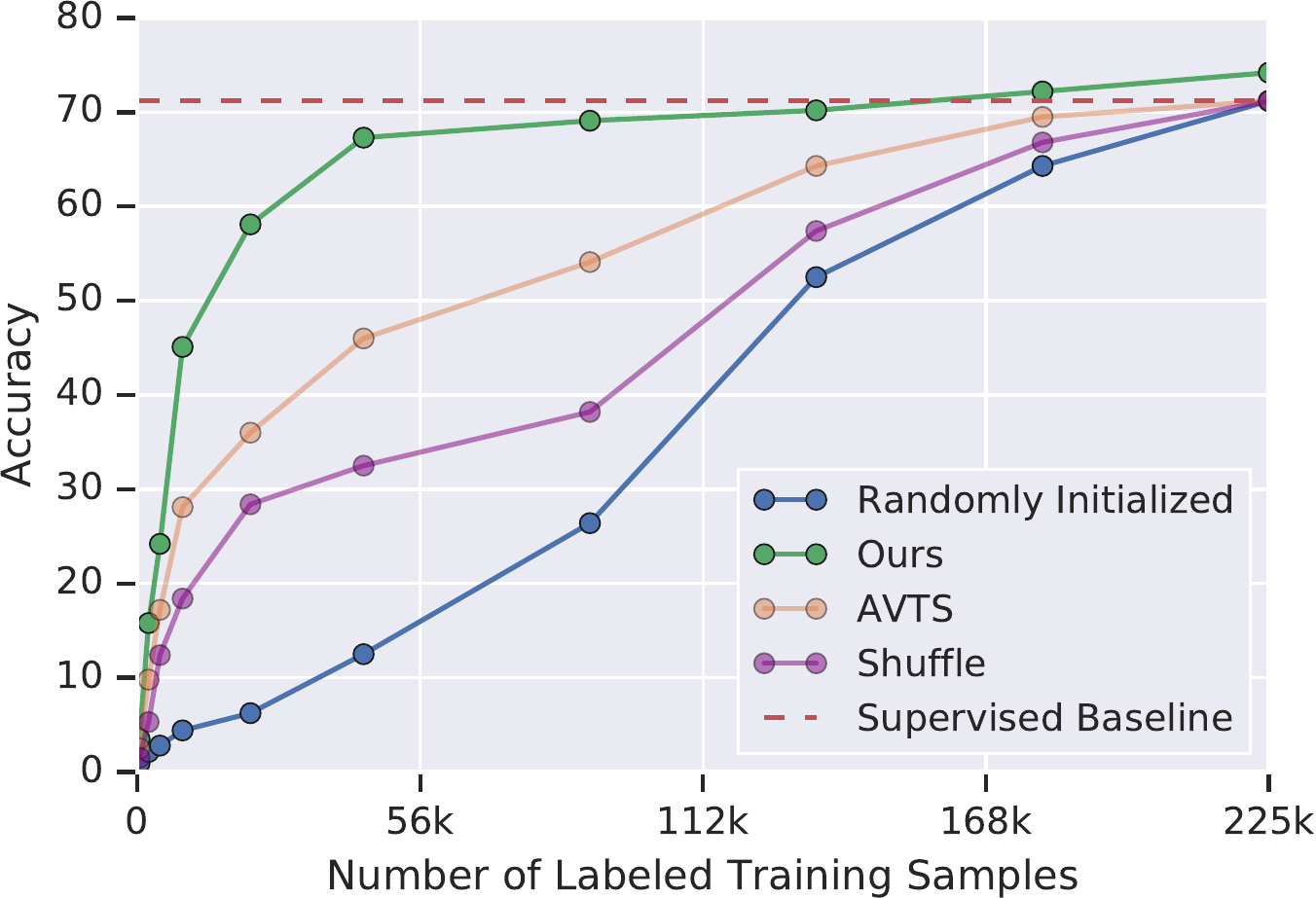}
    \caption{How much labeled, supervised data is needed once the unsupervised representation is learned. We achieve comparable performance with roughly half the data and outperform the supervised baselines when using the entire dataset.}
    \label{fig:amounts-of-labeled-data}
\end{figure}

%\subsection{How much labeled data is needed?}
%\subsection{Using less data for supervised learning}
\paragraph{Improving supervised learning}
Once we have learned a representation space using large amounts of unlabeled data, we want to determine how much labeled data is needed to achieve competitive performance. In Fig. \ref{fig:amounts-of-labeled-data}, we compare various approaches trained using our unlabeled videos then fine-tuned on Kinetics using different amounts of labeled data. The Kinetics dataset has 225k labeled samples and we find that using only 25k (10\%) yields reasonable performance (58.1\% accuracy), only 11\% lower than our baseline, fully-supervised model using all samples.

Further, we find we are able to match performance using only 120k samples, about half the dataset. Using the entire dataset, we outperform the baseline network, due to better initilizations and the distillation of various modalities into the RGB stream.

\begin{table}
    \centering
    \begin{tabular}{lcc}
    \toprule
    Method & HMDB & UCF101 \\
    \midrule
    \multicolumn{3}{l}{\textbf{Supervised}}\\
    3D ResNet-50 Scratch   & 35.2 & 63.1 \\
    3D ResNet-50 ImageNet & 49.8 & 84.5 \\
    3D ResNet-50 Kinetics & 74.3 & 95.1 \\
    \midrule
    \multicolumn{3}{l}{\textbf{Unsupervised}}\\
    Shuffle \cite{misra2016shuffle} & 18.1 & 50.2\\
    %O3N \cite{fernando2017self}  & 32.5 & 60.3 \\
    OPN \cite{lee2017unsupervised} & 37.5 & 37.5 \\
    %Patch \cite{wang2015unsupervised} & - & 41.5 \\
    %Multisensory \cite{owens2018audio} & - & 82.1 \\
    AVTS \cite{korbar2018cooperative} & 61.6 & 89.0 \\
    \midrule
    Our Evolved Loss & 66.2 & 92.4 \\
    \bottomrule
    \end{tabular}
    \caption{State-of-art comparisons on HMDB51 and UCF101. The top shows results for 3D ResNet-50 when pretrained using supervised datasets. Note that previous approaches train on videos from existing datasets (e.g., Kinetics), whereas we use random video clips. Even using more difficult data, we outperform the previous methods.}
    \label{tab:state-of-art}
\end{table}

\begin{figure}
    \centering
    \includegraphics[width=0.4\linewidth]{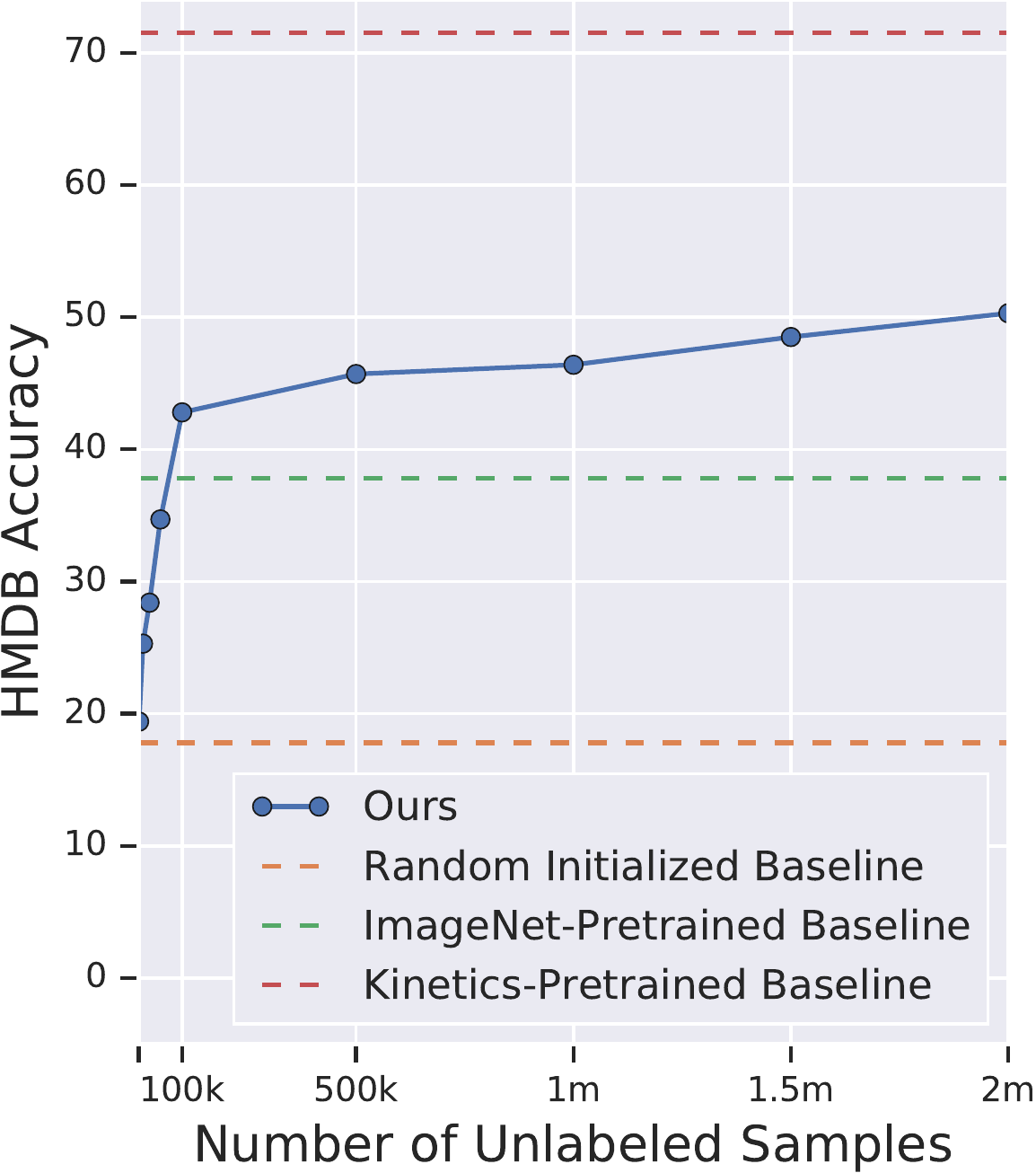} \includegraphics[width=0.4\linewidth]{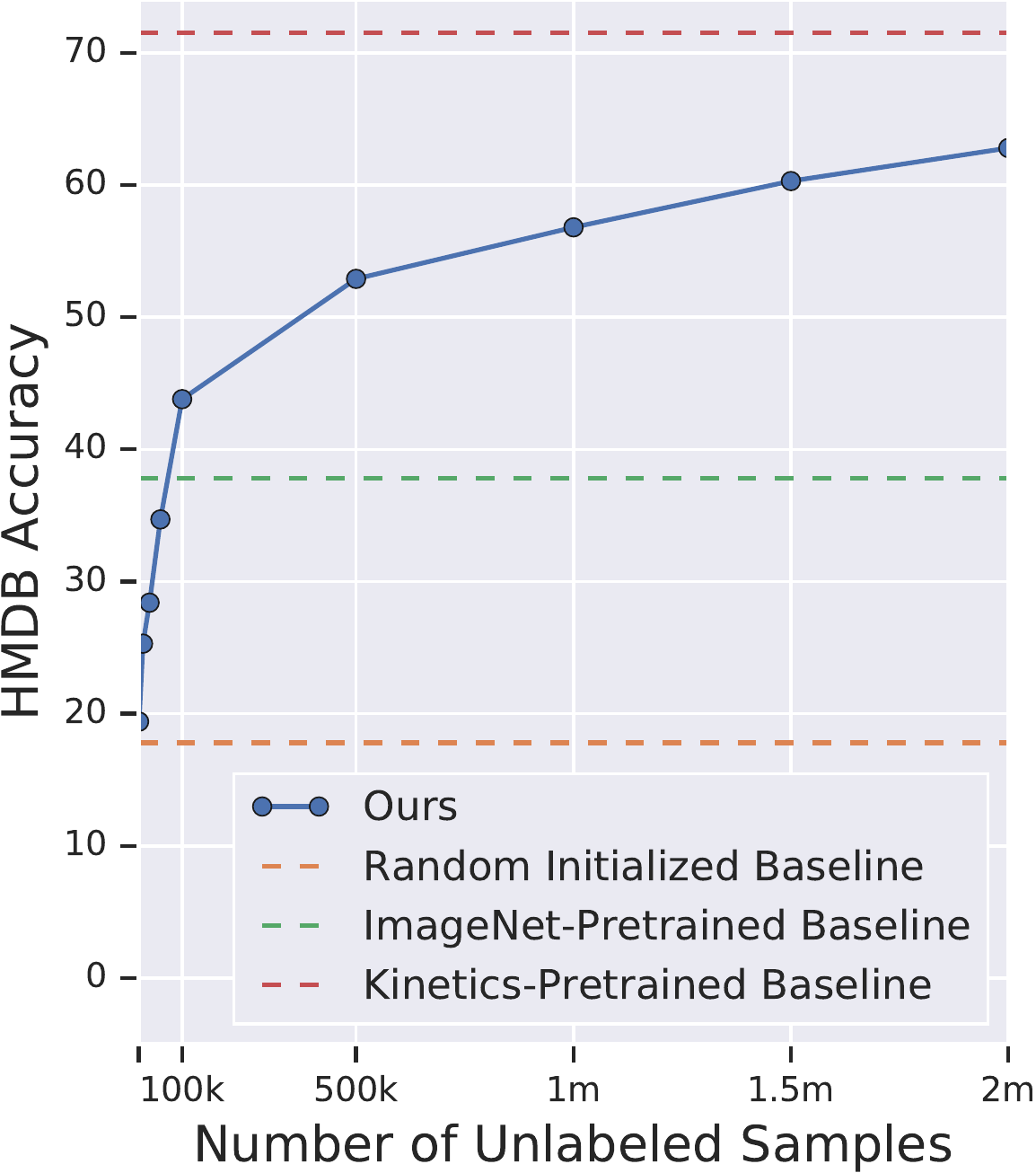}
    \caption{Comparisons of different amounts of unsupervised data. \textbf{Left}: Total number of training iterations fixed (i.e., less epochs as data is added). \textbf{Right}: Total number of epochs fixed (i.e., more iterations as more data is added). We observe that adding more data without increasing training time improves performance, while training longer on more data is better. On HMDB.}
    \label{fig:amounts-of-unsupervised-data}
\end{figure}

\paragraph{Comparison to previous methods}

In Table \ref{tab:hmdb-1}, we compare various self-supervised methods to our evolved loss function on our unlabeled videos. We find that while all approaches outperform the randomly initialized networks, only our evolved loss function outperforms ImageNet pretraining and performs comparably to the pretrained network with the labeled Kinetics data. We also confirm that evolving the loss function is beneficial by comparing to a random function.

In Table \ref{tab:state-of-art}, we compare our approach to previous reported methods. We find that even though our approach is using more difficult unlabeled data, we still outperform the exiting methods by a significant margin, including supervised training on HMDB and ImageNet data.
\vspace{-1em}

\begin{figure}
    \centering
    \includegraphics[width=\linewidth]{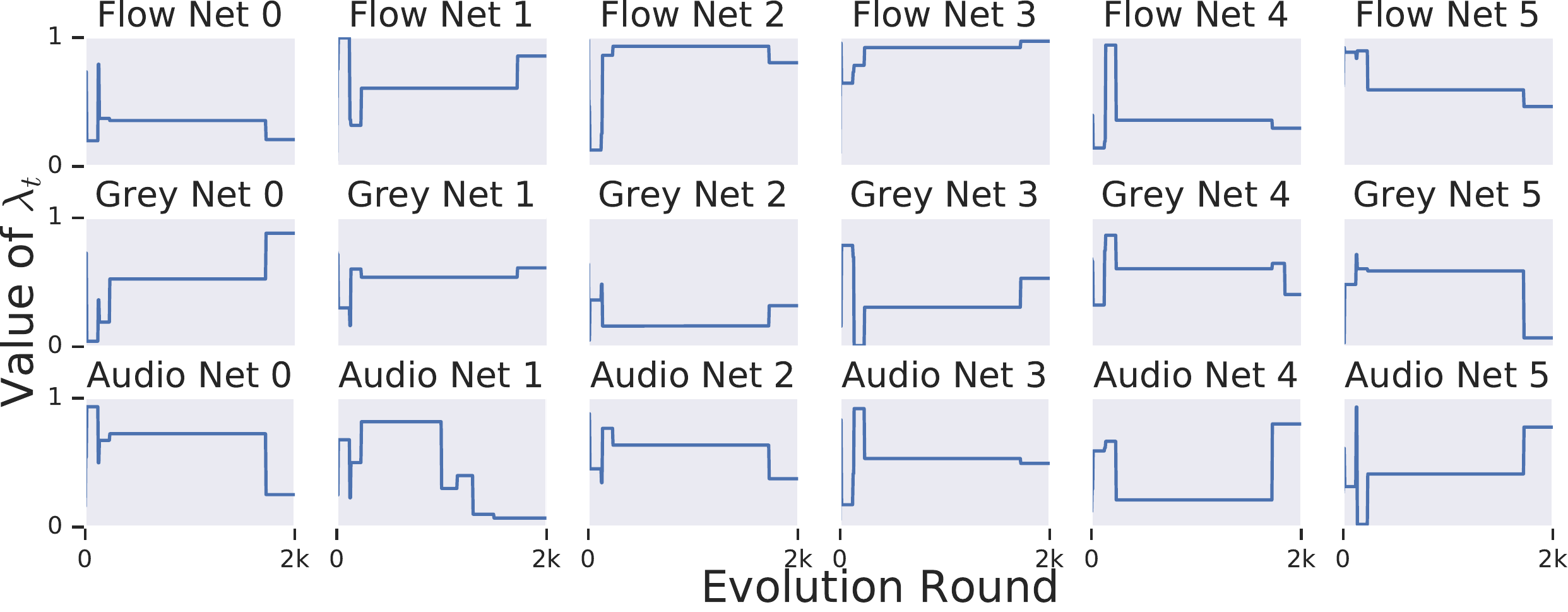}
    \caption{Values of the distillation weights during evolution. `The learned loss function distills audio and flow features later in the network (when features are more abstract), while grey-modality features are distill more early on (likely because they more easily match the RGB features).}
    \label{fig:distill-evol}
\end{figure}

\begin{figure}
    \centering
    \includegraphics[width=0.4\linewidth]{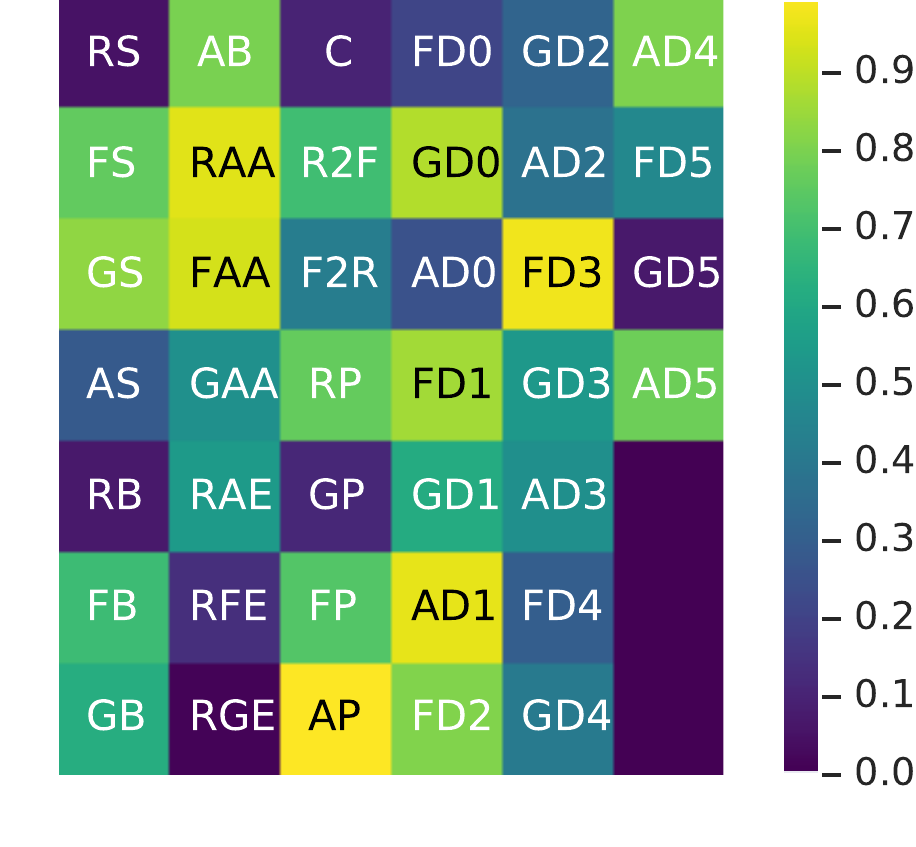}
    \caption{Heatmap visualization of the learned loss function. Higher values indicate the components importance. The first letter representation modality (R=RGB, A=Audio, F=Flow, G=Grey), The tasks are S=Shuffle, C=colorize, A=Audio align, P=Future prediction, B=backward detection, D=Distill, E=Embed. The numbers indicate the layer the distillation loss is applied.}
    \label{fig:loss-heatmap}
\end{figure}

\paragraph{Benefit of additional unlabeled data}
We compare different amounts of unlabeled data in Fig. \ref{fig:amounts-of-unsupervised-data} finding that using more data is always beneficial.

\vspace{-2mm}
\section{Evolved Loss Function Analysis}

Examining the weights of the evolved loss function, $\lambda_{m,t}$ and $\lambda_d$, allows us to check which tasks are more/less important for the target task. Fig. \ref{fig:tasks-evol} illustrates the weights for each task ($\lambda_{m,t}$) over the 2000 evolution rounds. We observe tasks such as RGB frame shuffle, colorization, etc. get very low weights, suggesting they are not very useful for the action recognition task. Tasks such as audio alignment, future frame prediction, and cross-modality reconstruction (e.g., RGB to flow) are quite important.

Fig. \ref{fig:distill-evol} shows the weights corresponding to various distillation losses for each modality. We find that distilling greyscale features is beneficial for early convolutional layers, likely because the representations are more similar to RGB. Audio and flow representations are distilled more strongly later in the network, when the features are more abstract. In Fig. \ref{fig:loss-fn-evol}, we show a heatmap representation of the weights during evolution and our final fully-evolved loss is shown in Fig. \ref{fig:loss-heatmap}.

{\footnotesize
\bibliographystyle{ieee_fullname}
\bibliography{egbib}
}

\end{document}